\documentclass{article}
\usepackage{spconf,amsmath,graphicx}

\usepackage[hidelinks]{hyperref}
\usepackage{booktabs}
\usepackage{subcaption}
\usepackage[dvipsnames]{xcolor}

\usepackage{pifont}
\newcommand{\cmark}{\ding{51}}%
\newcommand{\xmark}{\ding{55}}%

\newcommand{\bg}[1]{\boldsymbol{#1}} 
\newcommand{\bm}[1]{\mathbf{#1}} 

\newcommand\raiseT[2]{%
\setbox0\hbox{$#1{#2}$}\raise\dp0\box0}

\usepackage{parskip}
\usepackage{caption}
\title{RSUD20K: A Dataset for Road Scene Understanding In Autonomous Driving}
\name{Hasib Zunair, Shakib Khan, and A. Ben Hamza}
\address{Concordia University, Montreal, QC, Canada}
\begin{document}
%
\maketitle
\begin{abstract}
Road scene understanding is crucial in autonomous driving, enabling machines to perceive the visual environment. However, recent object detectors tailored for learning on datasets collected from certain geographical locations struggle to generalize across different locations. In this paper, we present RSUD20K, a new dataset for road scene understanding, comprised of over 20K high-resolution images from the driving perspective on Bangladesh roads, and includes 130K bounding box annotations for 13 objects. This challenging dataset encompasses diverse road scenes, narrow streets and highways, featuring objects from different viewpoints and scenes from crowded environments with densely cluttered objects and various weather conditions. Our work significantly improves upon previous efforts, providing detailed annotations and increased object complexity. We thoroughly examine the dataset, benchmarking various state-of-the-art object detectors and exploring large vision models as image annotators. Dataset, code and pre-trained models are available at \textcolor{blue}{https://github.com/hasibzunair/RSUD20K}
\end{abstract}
%

\section{Introduction}
Effectively understanding road scenes involves overcoming challenges related to illumination, viewpoint differences, and background clutter, which significantly impact object detection performance, especially in cases of occluded and small objects. Addressing these issues is crucial for deploying robust perception systems, ensuring safe driving of autonomous vehicles, and averting accidents in real-world scenarios. Recent approaches to road scene understanding are data-driven learning-based methods like YOLO~\cite{terven2023comprehensive}, Faster R-CNN~\cite{ren2015faster} and DETR~\cite{carion2020end}, and have become prevalent in modern autonomous driving systems. Publicly available datasets such as CityScapes~\cite{Cordts2016Cityscapes}, Mapillary Vistas~\cite{neuhold2017mapillary}, KITTI Vision Benchmark Suite~\cite{Geiger2012CVPR}, BDD100K~\cite{yu2020bdd100k}, and Waymo Open Dataset~\cite{sun2020scalability} have paved the path toward autonomous driving. However, learning-based methods trained on these datasets may not generalize well to new geographic locations such as Bangladesh or India. Since data labeling is both time consuming and costly, a workaround to this problem is to leverage the zero-shot capabilities of recent vision foundation models such as Grounding DINO~\cite{liu2023grounding}, OWL-ViT~\cite{minderer2205simple}, DETIC~\cite{zhou2022detecting} and SAM~\cite{kirillov2023segment} to quickly annotate new data using prompt engineering. While automatic data labeling could potentially save time and reduce cost, it is still questionable as to how good supervised models would perform when trained on data generated using foundation models.

\begin{figure*}[htb!]
\centering
\includegraphics[scale=.8]{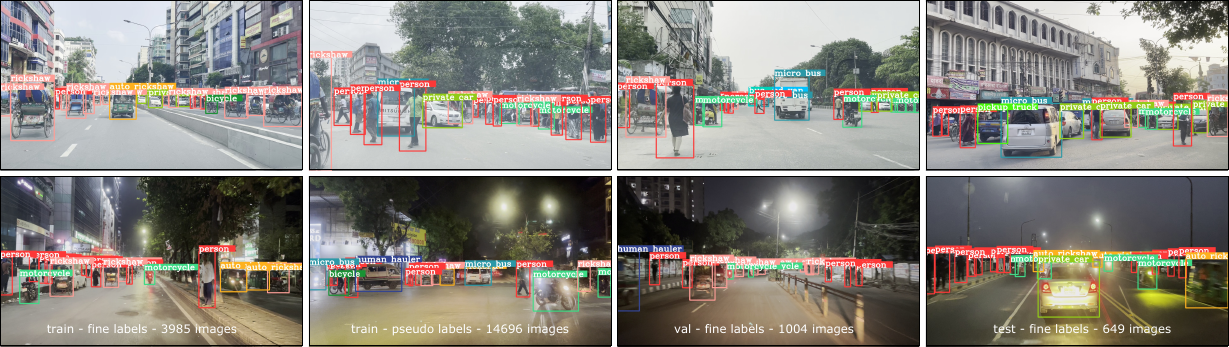}
\caption{\textbf{RSUD20K for road scene understanding}. The dataset consists of a total of \textbf{20334} images with \textbf{130K} bounding box annotations of \textbf{13} different objects. Images are captured from the driving perspective of diverse road scenes, objects from different viewpoints, occlusions, as well as under various weather conditions.}
\label{Fig:dataset}
\end{figure*}

To address the aforementioned issues and establish a platform to enable autonomous driving progress across various geographical locations, we propose a \textbf{R}oad \textbf{S}cene \textbf{U}nderstanding \textbf{D}ataset (RSUD20K) comprised of over 20K high-resolution images paired with 130K bounding box annotations of 13 objects captured from the perspective of a driving vehicle on Bangladesh roads. As shown in Fig.~\ref{Fig:dataset}, images consist of real-world scenarios like diverse roads, narrow streets, highways and objects from different viewpoints. Further, scenes include crowded environments and various weather conditions. The main contributions of this work can be summarized as follows: \textbf{(1)} We introduce RSUD20K, a novel dataset for road scene understanding, providing detailed annotations; \textbf{(2)} we benchmark state-of-the-art object detectors on RSUD20K, demonstrating the challenges in the task; and \textbf{(3)} we adopt a data-centric approach and assess the performance of large vision models as image annotators in zero-shot scenarios.


\section{Related Work}
\noindent\textbf{Bangladesh Street View Datasets.}\quad DhakaAI~\cite{DVN} presents a dataset with 3953 images and 24368 bounding box annotations across 21 classes. However, issues such as mislabeling and inconsistent image quality have been identified~\cite{rahman2022densely}. Poribohon-BD~\cite{tabassum2020poribohon} consists of 9058 images and bounding box labels for 15 vehicle classes. However, some objects such as boats are unrelated to autonomous driving and the dataset's multi-class vehicle category has limited samples. MVINet~\cite{ahmed2023deep} introduce a dataset for multi-class classification with 10 vehicle types and 100 images per class. This dataset is, however, not publicly available. Table~\ref{Tab:datasets_comparison} provides a detailed comparison between RSUD20K and these datasets.

\begin{table*}
\caption{\textbf{Comparison of RSUD20K with existing Bangladesh road scene understanding datasets}. While the other datasets are mainly for traffic detection and recognition of vehicles, RSUD20K is built for the application of autonomous driving.}
\centering
\begin{tabular}{lllll}
\toprule
Specifications & Dhaka-AI~\cite{DVN} & Poribohon-BD~\cite{tabassum2020poribohon} & MVINet~\cite{ahmed2023deep} & RSUD20K (Ours) \\
\midrule
Image Resolution & $1920\times1080$ & $1920\times1080$ & $128\times128$ & $1920\times1080$ \\
Number of Images & 3953 & 9058 & 17000 & 20000 \\
Number of Classes & 21 & 15 & 17 & 13\\
Target Task & Object Detection &  Classification & Multi-class  Classification & Object Detection \\
Annotation Type & 2D bounding boxes & 2D bounding boxes & 2D bounding boxes & 2D bounding boxes \\
Public & \cmark & \cmark & \xmark & \cmark\\
Application & Traffic Detection & Vehicle Classification & Vehicle Classification & Autonomous Driving \\
\bottomrule
\end{tabular}
\label{Tab:datasets_comparison}
\end{table*}

\smallskip\noindent\textbf{Object Detectors.}\quad Most existing object detection methods can be categorized into one- and two-stage detectors. One-stage detectors, such as YOLO~\cite{terven2023comprehensive}, single shot detector (SSD)\cite{liu2016ssd}, RetinaNet\cite{lin2017focal}, CenterNet~\cite{duan2019centernet}, and DETR~\cite{carion2020end}, directly generate candidate bounding boxes and predictions in a single processing step, making them suitable for real-time detection in autonomous driving. However, their performance notably degrades when detecting dense and small objects. On the other hand, two-stage detectors like R-CNN~\cite{girshick2014rich}, Fast R-CNN~\cite{girshick2015fast}, Faster R-CNN~\cite{ren2015faster}, and Mask R-CNN~\cite{he2017mask} employ object regions using deep features before classifying the objects, resulting in higher detection accuracy.

\smallskip\noindent\textbf{Large Vision Models (LVMs).}\quad The open-set capabilities of LVMs enable interpretation of objects unseen during training (i.e., zero-shot). Grounding DINO~\cite{liu2023grounding} is an open-set object detector based on the Transformer detector DINO, featuring grounded pre-training to detect arbitrary objects from category names or referring expressions. OWL-ViT transfers image-level contrastive pre-training to open-set object detection. DETIC~\cite{zhou2022detecting} extends open-set detection to over 20K classes by training classifiers on image-level data. SAM~\cite{kirillov2023segment} introduces a task and promptable model for image segmentation, trained on a dataset of 1B masks and 11M images. LVMs can be used for automatic data annotation (pseudo-labeling), potentially saving time and reducing cost. However, the performance of supervised models trained on datasets created using LVMs remains unclear.

\section{Road Scene Understanding Dataset}
\begin{figure*}[!htb]
\centering
\includegraphics[scale=0.8]{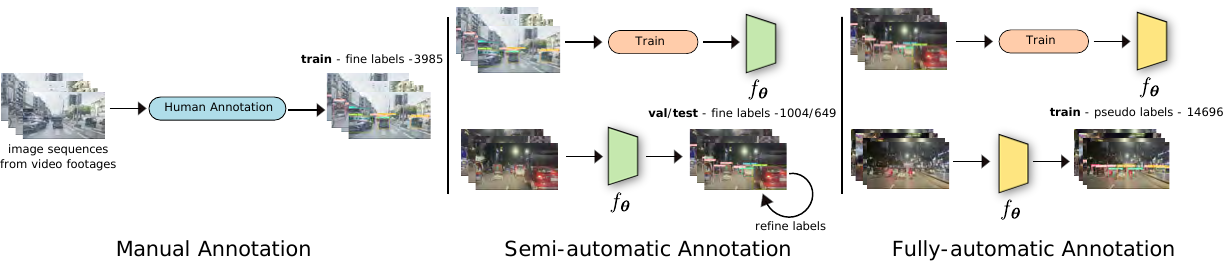}
\caption{\textbf{Overview of our data engine used to create RSUD20K}. We use a three-stage approach, where $f_{\boldsymbol{\theta}}$ is YOLOv6-M6 detector trained to predict bounding boxes and class labels of objects.}
\label{Fig:pipeline}
\end{figure*}

We start with the problem formulation, and then delve into the data engine and analysis of our RSUD20K dataset. The key steps of our data engine are presented in Fig.~\ref{Fig:pipeline}.

\smallskip\noindent\textbf{Problem Statement.}\quad Let $\mathcal{D}=\{(\bm{I}_i,\bm{B}_i)\}_{i=1}^{N}$ be a training set of $N$ labeled images $\bm{I}_{i}\in\mathcal{X}$ and their ground-truth bounding box annotations $\bm{B}_{i}$. Each bounding box represents the coordinates of the top-left and bottom-right corners of the bounding box in the image, denoted as $(x_{i}^{\text{min}}, y_{i}^{\text{min}}, x_{i}^{\text{max}}, y_{i}^{\text{max}})$. The task of object detection is to learn a model $f_{\bg{\theta}}: \mathcal{X}\to \mathcal{Y}$, where $\bg{\theta}$ is a set of learnable parameters. In this context, $\mathcal{Y}$ represents the set of bounding box annotations for the detected objects. Given a test image $\bm{I}$, the trained model predicts a set of bounding boxes $\bm{B}_{\text{p}}$, where each predicted bounding box is denoted as $(x_{\text{p}}^{\text{min}}, y_{\text{p}}^{\text{min}}, x_{\text{p}}^{\text{max}}, y_{\text{p}}^{\text{max}})$.

\subsection{Data Engine}
\noindent\textbf{Image Collection.}\quad We created our RSUD20K dataset by collecting a diverse set of over 80 video sequences with a resolution of $1920 \times 1080$ featuring street scenes, recorded using an in-car dashboard-mounted camera. To ensure data integrity and avoid leakage, we manually grouped the video sequences into training, validation and test sets, balancing scenes from different environments such as outskirts and cities, along with varying weather conditions like nighttime or rainy scenes. Each video sequence, lasting 2-5 minutes, showcases different roads, streets, and highways in the same area. The group partitioning resulted in 54 training, 19 validation, and 11 test video sequences. From these sequences, we sampled images for training, validation, and test sets, with 400, 60, and 60 frames respectively per video. This sampling strategy yielded a total of 23,246 images, with 18,762 training images, 1,008 validation images and 656 test images.

\smallskip\noindent\textbf{Manual Annotation.}\quad In this stage, we follow a process similar to traditional multi-class object detection. A team of professionals manually annotates 4,000 randomly sampled images from the training set by delineating bounding box regions around objects of interest and assigning class names using the open-source Label Studio annotation tool\footnote{https://labelstud.io}. The focus is on creating a dataset for autonomous driving, specifically for understanding driving scenarios. We identify 13 classes of interest: \textit{person, rickshaw, rickshaw van, auto rickshaw, truck, pickup truck, private car, motorcycle, bicycle, bus, micro bus, covered van, and human hauler}. Annotators are instructed to draw bounding boxes tightly around these objects, ensuring at least one bounding box with a class label per image. Images without objects from the specified 13 classes are discarded. If an object is occluded by more than roughly 50\%, it is also omitted from annotation. In cases where the object is visible, but still partially occluded, a box is drawn around the visible portion. The constraint to draw tight boxes is relaxed in dense and cluttered scenes to capture real-world characteristics. On average, the annotation time per image with multiple bounding boxes is 48 seconds. This stage yields 3,985 images with corresponding bounding box labels for a total of 13 different classes.

\smallskip\noindent\textbf{Semi-Automatic Annotation.}\quad In this stage, we aim to label the validation and test set image collections, consisting of 1,008 and 656 images, respectively. Initially, we train YOLOv6-M6 using the labeled data from manual annotation. Through inference, we generate bounding box predictions for the 13 classes within the validation and test sets in a pseudo-labeling fashion. Images with null predictions are excluded. Subsequently, we present the pseudo-labeled images to annotators for label refinement, which involves adjusting bounding boxes tightly around objects and correcting any inaccurate labels predicted by the model. This stage serves two main purposes: (i) to establish gold standard labels for the validation and test sets for model evaluation, and (ii) to expedite the labeling process by leveraging model predictions. The average annotation time per image is significantly reduced from 48 seconds to 8 seconds, representing a six-fold increase in efficiency compared to manual human annotation. This approach is accurate, efficient, cost-effective, and scalable, playing a crucial role in building high-quality datasets. Finally, we generate refined validation and test sets comprising 1,004 and 649 images, respectively, each with corresponding bounding box labels for the 13 classes. These sets, combined with the training set, form the RSUD5K dataset.

\smallskip\noindent\textbf{Fully Automatic Annotation.}\quad In this final stage of our annotation process, automation takes center stage, leveraging a semi-supervised approach~\cite{zunair2021star}. We employ YOLOv6-M6, previously trained on images with bounding box labels in RSUD5K to pseudo-label unlabeled images. The model's predictions for bounding boxes across the 13 classes are generated for the remaining 14,762 images. After filtering out images with null predictions, we obtain 14,696 pseudo-labeled images. These are seamlessly integrated into the RSUD5K training set, resulting in the expanded RSUD20K dataset, which now comprises a total of 20,334 images.

\subsection{Data Analysis}
\noindent\textbf{Overview of Images and Bounding Boxes.}\quad Each image in our RSUD20K dataset is annotated with bounding boxes and their class labels, featuring annotations with multiple instances at various levels of granularity. RSUD20K encapsulates diverse driving scenes captured from the perspective of a driving vehice, encompassing objects in different viewpoints and densely cluttered areas leading to significant occlusions. Notably, the dataset includes driving scenes at night and in rainy conditions, distinguishing it from existing autonomous driving datasets.

\begin{figure}[!htb]
\centering
\includegraphics[scale=.2]{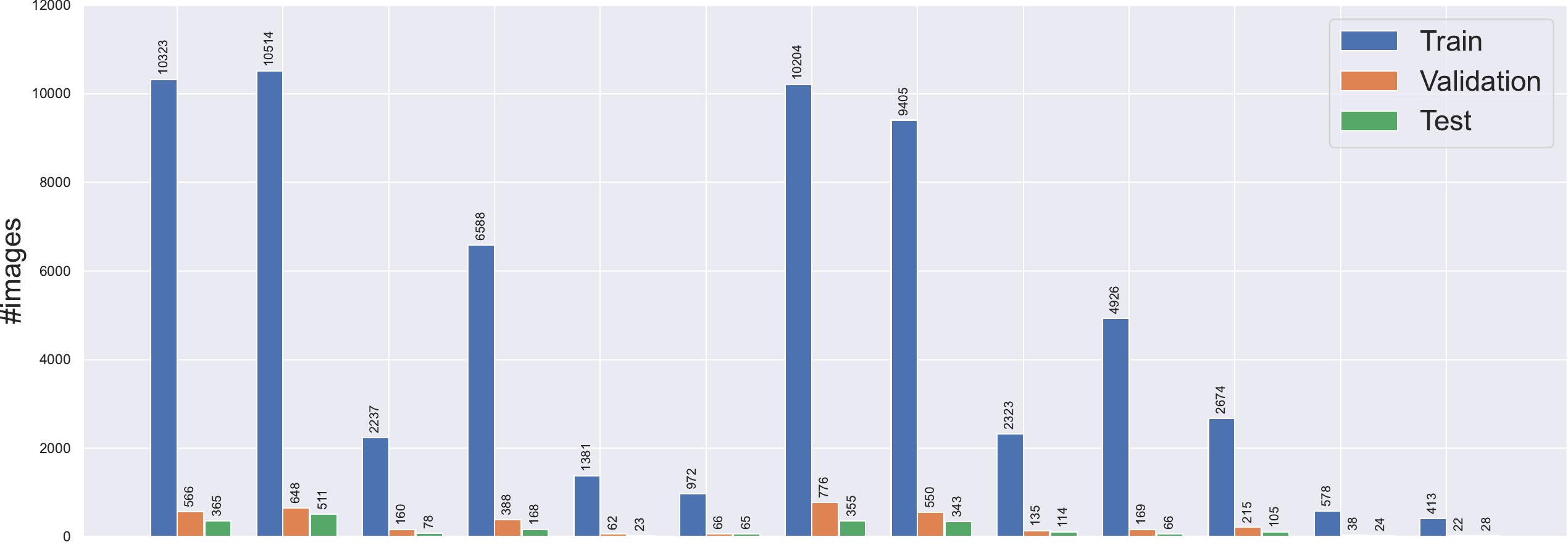}\\[0.1ex]
\includegraphics[scale=.2]{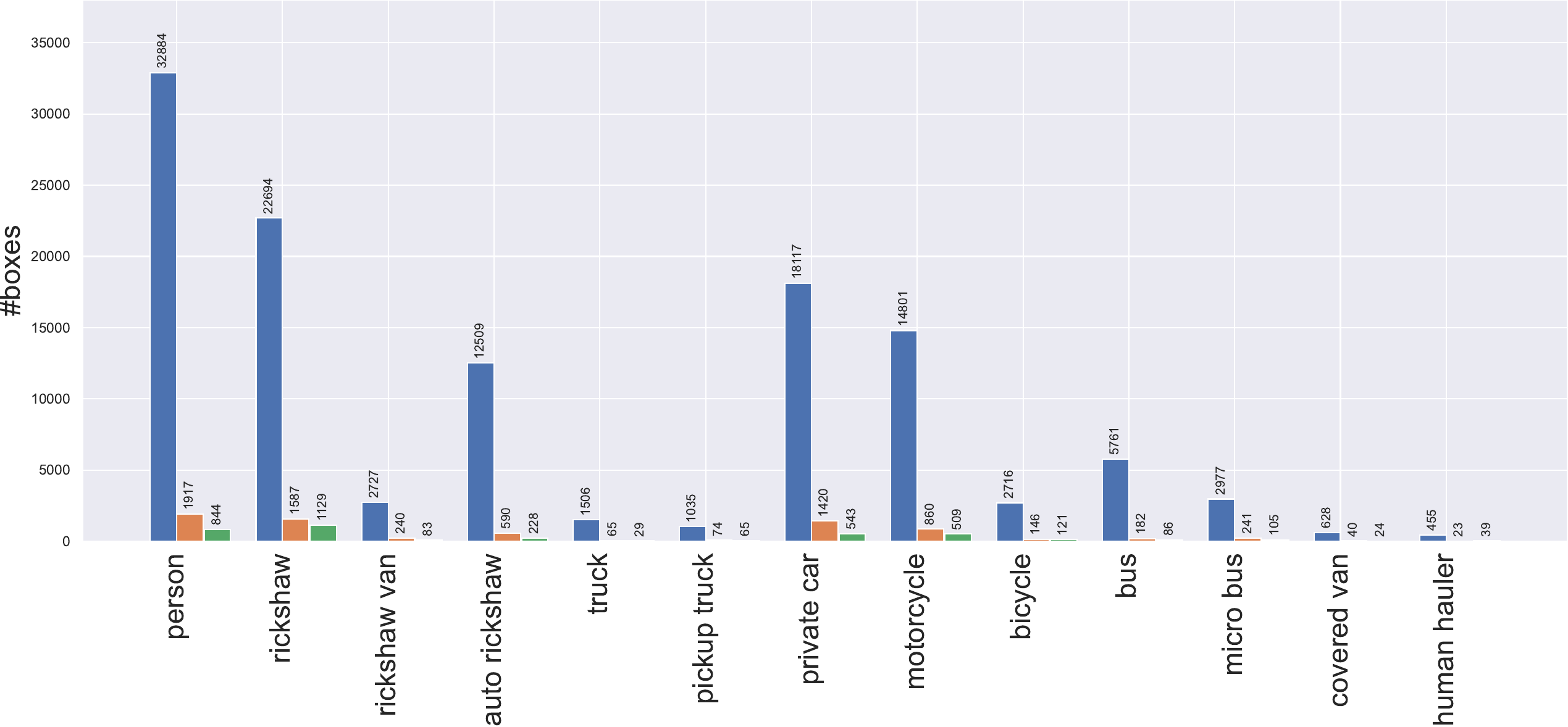}
\caption{\textbf{Dataset statistics of RSUD20K}. Image level annotations statistics (top) and bounding box instances (bottom).}
\label{Fig:stats}
\end{figure}

\smallskip\noindent\textbf{Annotation Statistics.}\quad We aim to answer two questions regarding class distribution: ``How many images in each dataset split belong to a particular class?'' and ``How many bounding box instances exist for each class?''. Fig.~\ref{Fig:stats} presents these statistics at both the image and instance levels. In Fig.~\ref{Fig:stats} (top), we depict the distribution of images containing different classes in all data splits. Notice the data imbalance, with over-represented classes like \textit{person, rickshaw, private car}, and under-represented classes such as \textit{pickup truck, covered van, human hauler}. Fig.~\ref{Fig:stats} (bottom) illustrates the distribution of bounding box instances for each class. The training set exhibits the highest number of instances for \textit{person} (over 30K boxes) and the lowest for \textit{human hauler} (455 instances), indicating significant imbalance. In summary, RSUD20K comprises 130K bounding box instances across 13 classes, distributed as 118,810, 7,385, and 3,805 instances in the training, validation, and test sets, respectively.

\smallskip\noindent\textbf{Object Coverage.}\quad Understanding objects encountered on the streets accurately is crucial for road scene comprehension and marks the initial step toward autonomous driving. Annotations include various object viewpoints (e.g., front, back, side), addressing the challenge posed by narrow, lane-less roads, where recognizing objects from different perspectives is vital. This distinguishes our dataset from existing ones. We also focused on establishing entry-level categories in lieu of fine-grained recognition, which involves distinguishing subordinate categories. For instance, our dataset includes fine-grained classes such as \textit{rickshaw} and \textit{rickshaw van}.

\smallskip\noindent\textbf{Scene Coverage.}\quad We collected diverse video sequences capturing various road/driving scenes, encompassing daylight, nighttime and rainy weather, marking distinctive features compared to existing datasets. Our dataset also includes scenes from heavily crowded areas and outskirts of cities, and  accounts for varying weather conditions (e.g., extreme sunny days causing glare), resulting in illumination challenges.

\begin{figure}[!htb]
\centering
\includegraphics[width=3.4in]{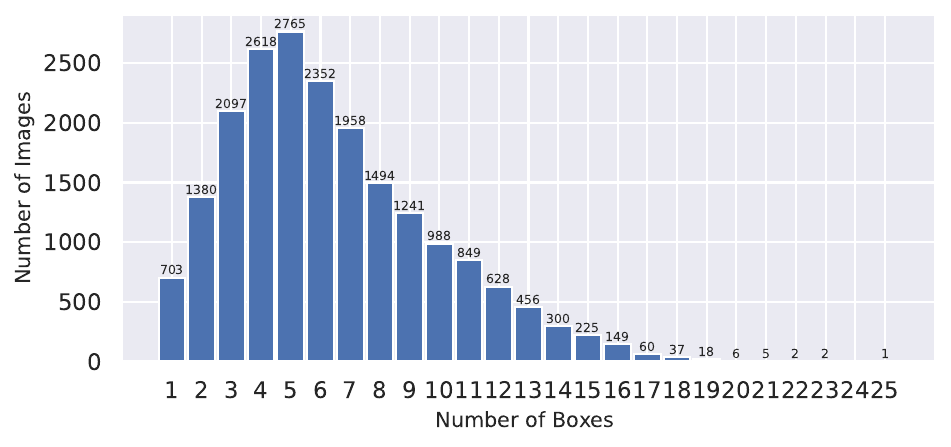}
\caption{\textbf{Distribution of box instances in RSUD20K}. Most images have a total of roughly 5 to 8 bounding boxes, demonstrating that there are many objects in the driving scenes.}
 \label{fig:bbox}
\end{figure}

\smallskip\noindent\textbf{Density of Bounding Boxes.}\quad In Fig.~\ref{fig:bbox}, we examine the distribution of bounding boxes in RSUD20K. On average, most images in the dataset contain around 4-9 objects. Over 3.5K images exhibit over 10 densely cluttered objects. The most frequent number of bounding boxes per image is 6, observed in 2765 images, followed by 2618 images with 5 bounding boxes and 2352 images with 7 bounding boxes. There are also 988 images with 10 bounding boxes and 628 images with 8 bounding boxes, and one image with 23 bounding boxes. This analysis provides valuable insights into the diversity and distribution of bounding box annotations in RSUD20K.

\smallskip\noindent\textbf{Challenges.}\quad RSUD20K presents several challenges: (i) It features scenes with no typical lane markings and narrow streets, which are uncommon in existing datasets; and (ii) certain vehicles like \textit{rickshaw} and \textit{rickshaw van} are human-powered, posing recognition difficulties as models may easily misclassify them as persons. Also, most images depict crowded street scenes with densely cluttered objects, leading to heavy occlusions. Certain scenes were captured both in daylight and nighttime, including rainy conditions, thereby introducing challenges related to varying illumination.

\section{Experiments}

\subsection{Experimental Setup}
\noindent\textbf{Baselines.}\quad We benchmark recent state-of-the-art object detectors on RSUD20K, including YOLOv6 and YOLOv8~\cite{terven2023comprehensive}, Transformer-based detectors DETR~\cite{carion2020end} and RTMDET~\cite{Lyu2022RTMDet}. We also compare the performance of supervised models when trained with labels generated by LVMs, including Grounding DINO~\cite{liu2023grounding}, OWL-ViT~\cite{minderer2205simple}, SAM~\cite{kirillov2023segment}, and DETIC~\cite{zhou2022detecting}.

\smallskip\noindent\textbf{Evaluation Metrics.}\quad We employ mean average precision (mAP), a standard metric in object detection, as our primary evaluation measure. Given the imbalanced nature of our dataset, we also report per-class mAP results.

\subsection{Benchmarking Results}
We report the detection performance of various object detectors on RSUD20K, and explore the zero-shot capabilities of LVMs as image annotators to train supervised models. 

\begin{figure*}[!htb]
\centering
\includegraphics[scale=.35]{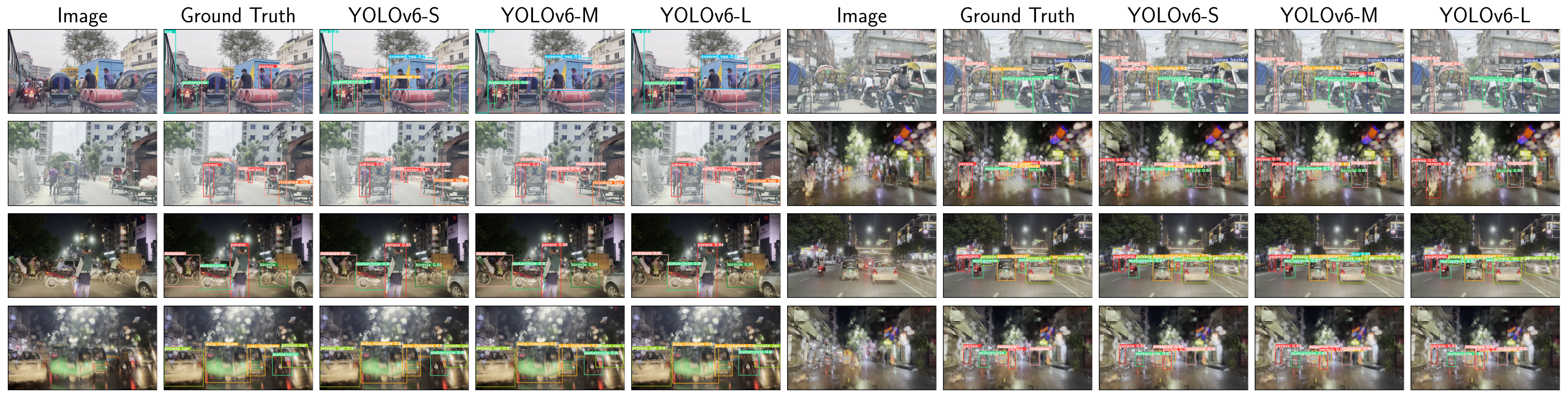}
\caption{\textbf{Visualization comparison of predictions by state-of-the-art object detectors on RSUD20K}. First two rows show occlusions and viewpoint differences. Second two rows show cases of various weather conditions. Zoom-in for better details.}
\label{Fig:qual}
\end{figure*}

\smallskip\noindent\textbf{Object Detection Results.}\quad We compare the performance of various state-of-the-art object detection methods, and the results are reported in Table~\ref{Tab:RSUD20K_sota}. We find that the best performing models are YOLOv6 and YOLOv8. Transformer-based methods like DETR yield lower detection performance, with 49.9 mAP compared to the best performing model YOLOv6-L achieving 73.7 mAP. Notably, objects operated by a person, such as rickshaws, rickshaw vans, and bicycles, showcase relatively efficient detection. However, certain classes like rickshaw van, truck, and pickup truck demonstrate lower mAPs (54.0, 50.7, and 67.9, respectively). This is because these object classes have sub-ordinate categories, resulting in size and shape variations, thus requiring fine-grained recognition. We also present qualitative examples in Fig.~\ref{Fig:qual}.

\begin{table*}[!htb]
\caption{\textbf{Performance comparison of state-of-the-art object detectors on RSUD20K in terms of mean and per-class mean average precision (mAP)}. The best results are in bold, and the second best are underlined.}
\centering
\resizebox{1\textwidth}{!}{
\begin{tabular}{lccccccccc}
        \toprule[1pt]
         Method & YOLOv8-L & YOLOv8-M & YOLOv8-S & RTMDET & DETR & YOLOv6-S & YOLOv6-M & YOLOv6-L\\
        \toprule[1pt]
        Input Size & $640\times 640$ & $640\times 640$ & $640\times 640$ & $640\times 640$ & $640\times 640$ & $640\times 640$ & $640\times 640$ & $640\times 640$\\
        Params (M) & 43.7 & 25.9 & 11.2 & 4.8 & 41.3 & 18.5 & 34.9 & 59.6\\
        \midrule[.8pt]
        person & \textbf{73.6} & 71.7 & 70.2 & 63.5 & 59.2 & 69.7 & \underline{72.8} & 72.5\\
        rickshaw & \textbf{88.3} & \underline{87.3} & 86.0 & 81.4 & 78.4 & 85.2 & 86.6 & 86.6 \\
        rickshaw van & 47.4 & \underline{53.8} & 52.8 & 49.8 & 33.4 & 45.9 & 51.5 & \textbf{54.0} \\
        auto rickshaw & \textbf{88.5} & 88.3 & 87.2 & 83.2 & 80.2 & 85.9 & 88.7 & \underline{88.4}\\
        truck & 34.8 & 40.7 & \underline{50.0} & 36.5 & 17.7 & 57.1 & 42.3 & \textbf{50.7} \\
        pickup truck & \textbf{67.9} & \underline{63.1} & 59.4 & 56.7 & 13.6 & 62.6 & 62.7 & 62.3\\
        private car & \textbf{87.9} & 87.6 & 86.5 & 81.1 & 77.8 & 85.9 & 87.2 & \underline{87.7}\\
        motorcycle & 77.1 & 76.9 & 75.1 & 68.9 & 67.1 & 75.2 & \underline{77.5} & \textbf{77.8}\\
        bicycle & 73.1 & 71.0 & 69.4 & 64.5 & 56.1 & 69.9 & \underline{73.9} & \textbf{74.5}\\
        bus & 64.1 & 60.6 & 65.1 & 60.8 & 41.5 & 70 & \underline{73.8} & \textbf{75.3}\\
        micro bus & 86.7 & \underline{87.1} & 76.8 & 81.0 & 48.8 & 86.0 & \textbf{88.1} & 85.9 \\
        covered van & 53.5 & 69.6 & 53.5 & 62.0 & 21.9 & 71.5 & \underline{71.6} & \textbf{78}\\
        human hauler & 67.1 & 75.3 & 70.7 & 60.8 & 57.0 & 71.4 & \underline{78.9} & \textbf{79.0}\\
        \midrule[.8pt]
        Mean & 70.4 & 71.8 & 69.4 & 65.4 & 49.9 & 72.0 & \underline{73.5} & \textbf{73.7}\\
        \bottomrule[1pt]
\end{tabular}} \label{Tab:RSUD20K_sota}
\end{table*}

\smallskip\noindent\textbf{LVMs as Image Annotators.}\quad We explore the use of LVMs for automatic image annotation to train supervised models, comparing their efficacy against human labels. Our data-centric approach keeps the algorithm fixed while altering the dataset, specifically the bounding box labels. We evaluate LVMs for data generation, aiming to determine if they can efficiently annotate data without compromising the detection performance. To automatically generate labels using LVMs, we first create a dictionary of class labels and descriptions by prompting a large language model to describe each class. Then, we input this dictionary and images from the training set of RSUD5K into various LVMs, including Grounding DINO, OWL-ViT, SAM and DETIC, which produce bounding box labels in a few hours. Finally, we train different object detection models on these LVM-generated datasets and compare their results on the test set (see Table~\ref{Tab:lvm_compare}). Our findings reveal that while LVMs efficiently annotate images for specialized tasks, the performance significantly drops across all object detection models compared to ground-truth labels. This is attributed to the distinctive nature of our dataset, featuring objects that are not commonly seen in other geographical locations (e.g., North America or Europe). Even DETIC, an object detector trained to detect 20K classes, exhibits suboptimal results in this specific context.

\begin{table}[!htb]
\caption{Performance comparison of large vision models trained on RSUD5K in terms of \%mAP.}
\small
\setlength\tabcolsep{5pt} 
\centering
\begin{tabular}{lccc}
\toprule
Method & YOLOv6-S & YOLOv6-M & YOLOv6-L\\
\midrule
Grounding DINO~\cite{liu2023grounding} & 9.0 & 8.8 & 9.3\\
OWL-ViT~\cite{minderer2205simple} & 10.1 & 9.9 & 10.7\\
SAM~\cite{kirillov2023segment} & 12.3 & 11.8 & 13.2\\
DETIC~\cite{zhou2022detecting} & \underline{14.8} & \underline{14.4} & \underline{15.8}\\
\midrule
Supervised & \textbf{66.3} & \textbf{65.7} & \textbf{67.2}\\
\bottomrule
\end{tabular}
\label{Tab:lvm_compare}
\end{table}

\subsection{Ablation Study}

We analyze how pseudo-labeled data affects model performance, and quantify the impact of increasing model capacity.

\begin{figure}[!htb]
     \centering
     \begin{subfigure}[b]{0.23\textwidth}
        \includegraphics[width=1\linewidth]{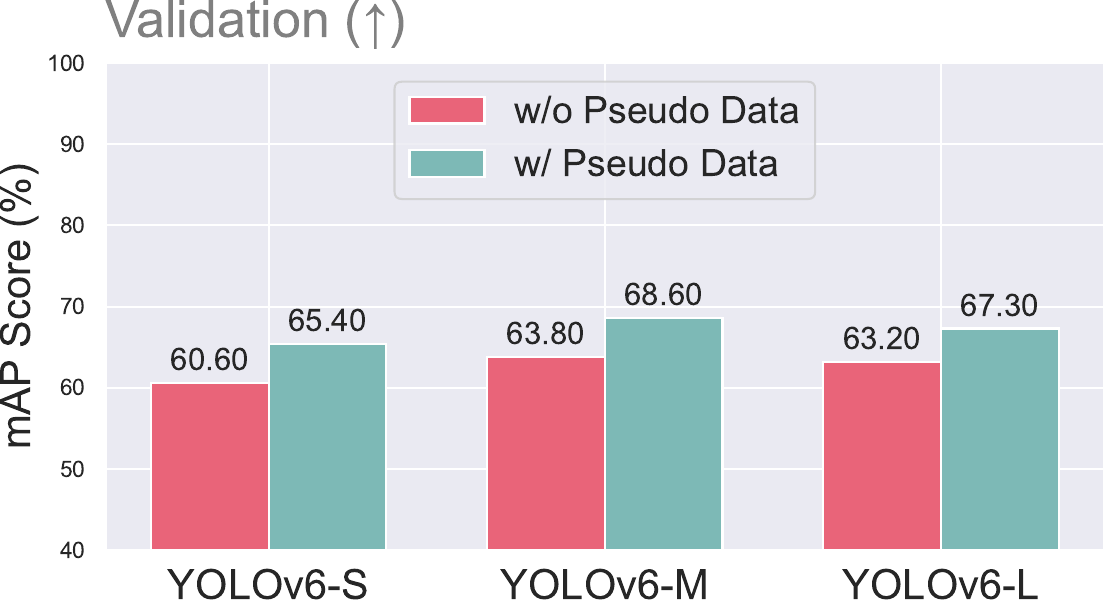}
        \caption{mAP on RSUD20K val set.}
        \label{fig:Ng1}
     \end{subfigure}
     \hspace{0.1cm}
     \begin{subfigure}[b]{0.23\textwidth}
        \includegraphics[width=1\linewidth]{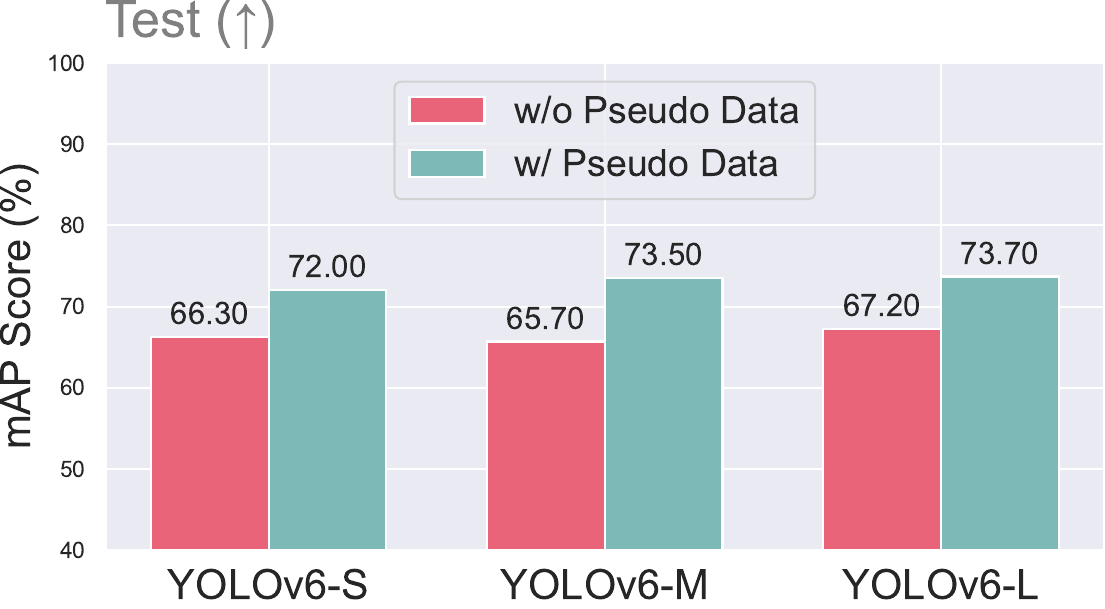}
        \caption{mAP on RSUD20K test set.}
        \label{fig:Ng2}
     \end{subfigure}
     \caption{\textbf{Performance comparison of models trained using pseudo labeled data on RSUD20K validation and test sets.} Pseudo-labels consistently improve detection performance different object detection models.}
     \label{fig:pseudo}
  \end{figure}

\smallskip\noindent\textbf{Data Scaling.}\quad Fig.~\ref{fig:pseudo} shows the benefit of using pseudo labeled data we created in the fully-automatic stage, resulting in consistently improved detection performance across various object detectors in both validation and test sets.

\smallskip\noindent\textbf{Model Scaling.}\quad We conduct experiments to investigate the impact of increasing model capacity on RSUD20K. As shown in Table~\ref{Tab:model_scaling}, the detection performance increases as the the model is scaled up, in both validation and test sets.

  \begin{table}[!htb]
     \caption{\textbf{Model scaling performance on RSUD20K}. Increasing model capacity yields higher detection performance.}
    \footnotesize
    \setlength\tabcolsep{1.3pt} 
     \centering
     \begin{tabular}{lcccc}
     \toprule[1pt]
     Method & Params (M) & FLOPs (G) & Validation (\%mAP) & Test (\%mAP)\\
     \midrule[.8pt]
     YOLOv6-S & 18.5 & 45.3 & 65.4 & 72.0 \\
     YOLOv6-M & 34.9 & 85.8 &  68.6 & 73.5 \\
     YOLOv6-L & 59.6 & 150.7 & 67.3 & 73.7 \\
     YOLOv6-M6 & 79.6 & 379.5 & \textbf{70.3} & \textbf{77.9} \\
     \bottomrule[1pt]
     \end{tabular}
     \label{Tab:model_scaling}
  \end{table}

\section{Conclusion}
We introduced RSUD20K, a novel dataset for road scene understanding, featuring 20K high-resolution images and 130K bounding box annotations across 13 object classes. These images capture diverse viewpoints, crowded environments, and under various weather conditions, presenting real-world scenarios. The significance of RSUD20K is underscored by its challenging nature, which remains unresolved even with state-of-the-art object detectors. The dataset not only serves as a benchmark for addressing existing vision challenges, but also provides a platform for developing road scene understanding algorithms tailored for autonomous driving.

\bibliographystyle{IEEEbib}
\bibliography{References}

\clearpage
\setcounter{page}{1}
\section*{----- Supplementary Material -----}
This supplementary material provides further details on implementation, along with additional experimental results.

\section{Implementation Details}
\noindent\textbf{Data Processing.}\quad Images are resized to $640 \times 640$ and standardized to have values between 0 and 1. We only apply basic data augmentation methods such as random flipping and resizing crops.

\medskip\noindent\textbf{Model Training.}\quad We used various open-source libraries to train different models. YOLOv6 models are trained using the out-of-the-box open-source library YOLOv6\footnote{\href{https://github.com/meituan/YOLOv6}{YOLOv6}}. YOLOv8 models are trained with the Ultralytics\footnote{\href{https://github.com/ultralytics/ultralytics}{ultralytics}}. DETR models utilize the MMDetection\footnote{\href{https://github.com/open-mmlab/mmdetection}{MMDetection}}. For RTMDET, we employ the MMYOLO\footnote{\href{https://github.com/open-mmlab/mmyolo}{MMYOLO}} library from OpenMMLab\footnote{\href{https://github.com/open-mmlab}{OpenMMLab}}. To train supervised models on labels generated by large vision models (LVMs), we used autodistill:  \textcolor{blue}{https://github.com/autodistill/autodistill}.

\medskip\noindent\textbf{Model Testing.}\quad After training, the model, when given an image as input, predicts bounding boxes and assigns labels from predefined classes.

\medskip\noindent\textbf{Hardware and Software Details.}\quad We conducted our experiments on a Linux workstation with a processing speed of 4.8Hz and 64GB RAM, featuring a single NVIDIA RTX 3080Ti GPU with 12GB of memory. All algorithms are implemented in PyTorch.

\medskip\noindent\textbf{Labeling Criteria.}\quad The data labeling criteria are as follows:
\begin{itemize}
  \item \textit{person}: draw boxes on persons only that are walking, not on vehicles.
  \item \textit{rickshaw}: boxes without person if possible. should be a tight box around the object.
  \item \textit{rickshaw van}: boxes around any three wheeler vans pulled by humans (e.g. selling vegetables or fruits).
  \item \textit{auto rickshaw}: any CNG, three wheeler electric vehicles
  \item \textit{truck}: big or small trucks
  \item \textit{pickup truck}: blue small vans, other small vans.
  \item \textit{private car}: any private car (includes jeeps too).
  \item \textit{motorcycle}: box should not have person if possible.
  \item \textit{bicycle}: box should not have person if possible.
  \item \textit{bus}: any bus, small or big.
  \item \textit{micro bus}: big cars like ambulance or other 7/8 seater cars (also Noah).
  \item \textit{covered van}: like pickup, but covered.
  \item \textit{human hauler}: leguna.
\end{itemize}

In general, we aim for tight boxes around objects. If an object is occluded by more than 50\%, we refrain from labeling it. We only draw a tight box if more than 50\% of the object is visible. In densely populated scenes, a bit of overlap between boxes is acceptable. Our data labeling process is facilitated using the Label Studio\footnote{\href{https://github.com/HumanSignal/labelImg}{Label Studio}} annotation tool.

\begin{figure}[!htb]
\centering
\includegraphics[width=3.45in]{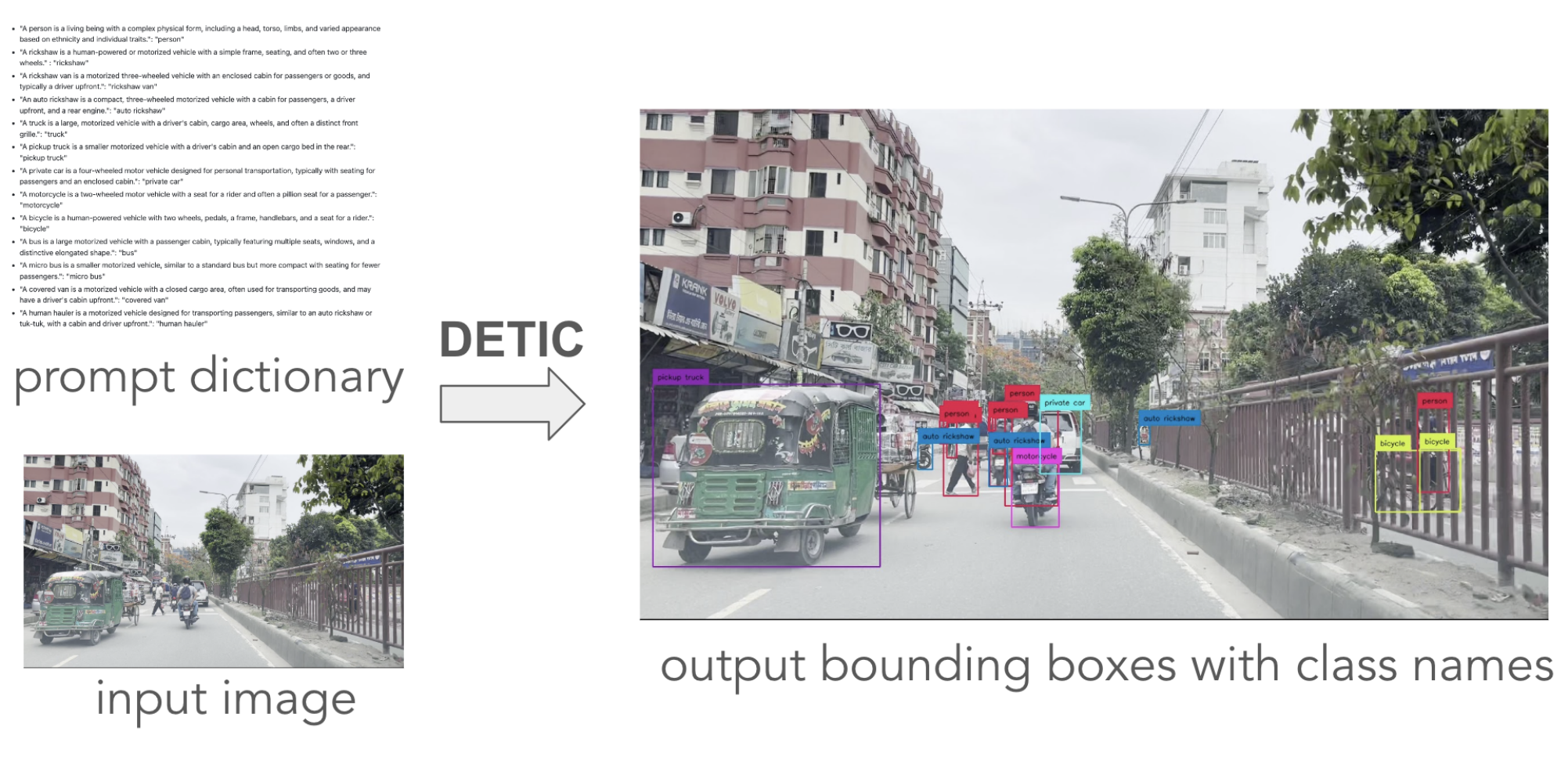}
\caption{\textbf{Label images via prompt engineering}. Given a prompt dictionary and an input image, a large vision model (i.e., DETIC) generates pseudo bounding box and class labels for our task.}
\label{fig:lvm_prompt}
\end{figure}

\medskip\noindent\textbf{Prompt Dictionary.}\quad We use the open-source language model, ChatGPT, to create short descriptions for our 13 objects. These descriptions form a prompt dictionary, guiding how large vision models are instructed. The dictionary includes text captions, serving as descriptions, and the corresponding class names. It specifies the actual labels. ChatGPT is employed to generate these label descriptions using the input prompt: ``describe what a {class name} looks like in 15-20 words.'' The generated descriptions are as follows:

\begin{itemize}
\item ``A person is a living being with a complex physical form, including a head, torso, limbs, and varied appearance based on ethnicity and individual traits.'' : ``person''
\item ``A rickshaw is a human-powered or motorized vehicle with a simple frame, seating, and often two or three wheels.'' : ``rickshaw''
\item ``A rickshaw van is a motorized three-wheeled vehicle with an enclosed cabin for passengers or goods, and typically a driver upfront.'' : ``rickshaw van''
\item ``An auto rickshaw is a compact, three-wheeled motorized vehicle with a cabin for passengers, a driver upfront, and a rear engine.'' : ``auto rickshaw''
\item ``A truck is a large, motorized vehicle with a driver's cabin, cargo area, wheels, and often a distinct front grille.'' : ``truck''
\item ``A pickup truck is a smaller motorized vehicle with a driver's cabin and an open cargo bed in the rear.'' : ``pickup truck''
\item ``A private car is a four-wheeled motor vehicle designed for personal transportation, typically with seating for passengers and an enclosed cabin.'' : ``private car''
\item ``A motorcycle is a two-wheeled motor vehicle with a seat for a rider and often a pillion seat for a passenger.'' : ``motorcycle''
\item ``A bicycle is a human-powered vehicle with two wheels, pedals, a frame, handlebars, and a seat for a rider.'' : ``bicycle''
\item ``A bus is a large motorized vehicle with a passenger cabin, typically featuring multiple seats, windows, and a distinctive elongated shape.'' : ``bus''
\item ``A micro bus is a smaller motorized vehicle, similar to a standard bus but more compact with seating for fewer passengers.'' : ``micro bus''
\item ``A covered van is a motorized vehicle with a closed cargo area, often used for transporting goods, and may have a driver's cabin upfront.'' : ``covered van''
\item ``A human hauler is a motorized vehicle designed for transporting passengers, similar to an auto rickshaw or tuk-tuk, with a cabin and driver upfront.'' : ``human hauler''
\end{itemize}
These descriptions serve as input prompts for foundation models such as DETIC, Grounding DINO, OWL-ViT, and SAM. Fig.~\ref{fig:lvm_prompt} illustrates the image labeling process via prompt engineering, where the pseudo bounding box and class labels are generated using DETIC as an image annotator.

\section{Limitations}
While RSUD20K aims to facilitate the development of road scene understanding algorithms tailored for autonomous driving, it does has certain limitations that warrant consideration. The dataset does not extensively address the fine-grained recognition of objects commonly encountered in everyday driving scenarios. For instance, a micro bus might serve as a police vehicle or even operate as an ambulance. Similarly, trucks come in various shapes and sizes, each serving a distinct purpose. While our work establishes a foundational set of objects for autonomous driving, a potential next step is to identify and collect images for these diverse use-cases.

\section{Additional Results}
\begin{figure*}[!htb]
\centering
\includegraphics[scale=.35]{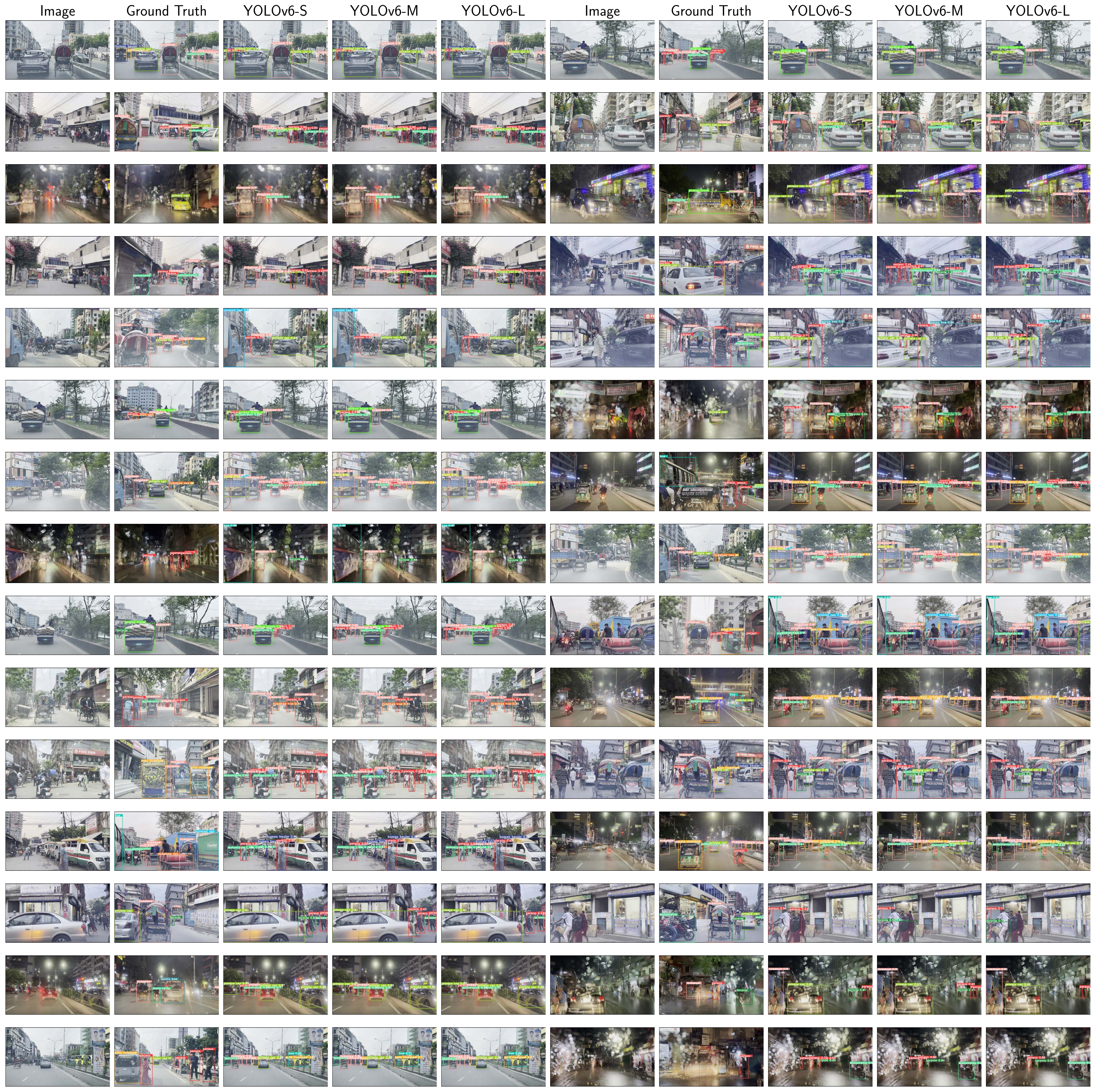}
\caption{\textbf{Visualization comparison of predictions by state-of-the-art object detectors on RSUD20K}. Zoom-in for better details.}
\label{Fig:compare_sota_ext}
\end{figure*}

\begin{figure*}[!htb]
\centering
\includegraphics[scale=.35]{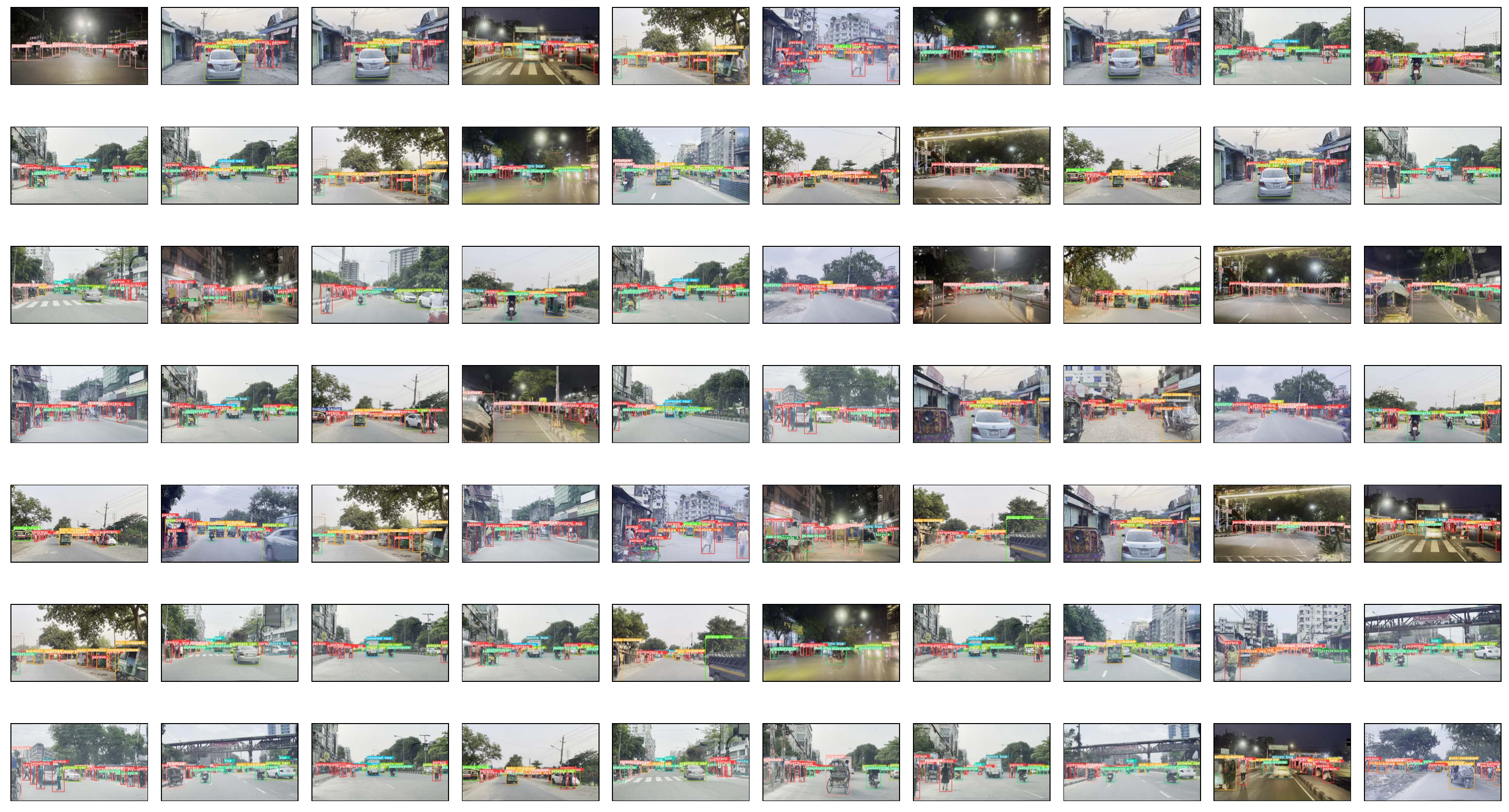}
\caption{\textbf{Randomly selected instances from RSUD20K dataset for road scene understanding in Bangladesh}. The dataset consists of a total of \textbf{20334} images with \textbf{130K} bounding box annotations of \textbf{13} different objects. Images are captured from the driving perspective of diverse road scenes, objects from different viewpoints, occlusions, as well as under various weather conditions.}
\label{Fig:samples}
\end{figure*}

In Figs.~\ref{Fig:compare_sota_ext} and \ref{Fig:samples}, we present additional experimental results on RSUD20K, showcasing model predictions and examples of driving scenes on Bangladesh roads.

\end{document}